\documentclass[biblatex,english]{lni}
\usepackage{booktabs}

\usepackage[]{blindtext}
\usepackage{dirtytalk}
\usepackage{url}

\usepackage[many]{tcolorbox}
\usepackage{xcolor}
\usepackage{varwidth}
\usepackage{environ}
\usepackage{xparse}
\usepackage{caption}
\usepackage{subcaption}

\newcommand{\rulesep}{\unskip\ \vrule depth -4em\ }

\newlength{\bubblesep}
\newlength{\bubblewidth}
\AtBeginDocument{\setlength{\bubblewidth}{.4\textwidth}}
\definecolor{bubbleprompt}{RGB}{0,150,255}  
\definecolor{bubbleresponse}{RGB}{241,240,240}

\newcommand{\bubble}[4]{%
  \tcbox[
    on line,
    arc=2mm,
    colback=#1,
    colframe=#1,
    #2,
    top=0.25pt,
    bottom=0.25pt,
    right=0.25pt,
    left=0.25pt
  ]{\color{#3}\begin{varwidth}{\bubblewidth}#4\end{varwidth}}%
}

\ExplSyntaxOn
\seq_new:N \l__ooker_bubbles_seq
\tl_new:N \l__ooker_bubbles_first_tl
\tl_new:N \l__ooker_bubbles_last_tl

\NewEnviron{prompt}
 {
  \begin{flushleft}
  \sffamily
  \seq_set_split:NnV \l__ooker_bubbles_seq { \par } \BODY
  \int_compare:nTF { \seq_count:N \l__ooker_bubbles_seq < 2 }
   {
    \bubble{bubbleprompt}{rounded~corners}{white}{\BODY}\par
   }
   {
    \seq_pop_left:NN \l__ooker_bubbles_seq \l__ooker_bubbles_first_tl
    \seq_pop_right:NN \l__ooker_bubbles_seq \l__ooker_bubbles_last_tl
    \bubble{bubbleprompt}{sharp~corners=southeast}{white}{\l__ooker_bubbles_first_tl}
    \par\nointerlineskip
    \addvspace{\bubblesep}
    \seq_map_inline:Nn \l__ooker_bubbles_seq
     {
      \bubble{bubbleprompt}{sharp~corners=east}{white}{##1}
      \par\nointerlineskip
      \addvspace{\bubblesep}
     }
    \bubble{bubbleprompt}{sharp~corners=northeast}{white}{\l__ooker_bubbles_last_tl}
    \par
   }
   \end{flushleft}
 }
\NewEnviron{response}
 {
  \begin{flushright}
  \sffamily
  \seq_set_split:NnV \l__ooker_bubbles_seq { \par } \BODY
  \int_compare:nTF { \seq_count:N \l__ooker_bubbles_seq < 2 }
   {
    \bubble{bubbleresponse}{rounded~corners}{black}{\BODY}\par
   }
   {
    \seq_pop_left:NN \l__ooker_bubbles_seq \l__ooker_bubbles_first_tl
    \seq_pop_right:NN \l__ooker_bubbles_seq \l__ooker_bubbles_last_tl
    \bubble{bubbleresponse}{sharp~corners=southwest}{black}{\l__ooker_bubbles_first_tl}
    \par\nointerlineskip
    \addvspace{\bubblesep}
    \seq_map_inline:Nn \l__ooker_bubbles_seq
     {
      \bubble{bubbleresponse}{sharp~corners=west}{black}{##1}
      \par\nointerlineskip
      \addvspace{\bubblesep}
     }
    \bubble{bubbleresponse}{sharp~corners=northwest}{black}{\l__ooker_bubbles_last_tl}\par
   }
  \end{flushright}
 }
\ExplSyntaxOff

\begin{document}
\title[Analyzing Chat Protocols of Novice Programmers]{Analyzing Chat Protocols of Novice Programmers Solving Introductory Programming Tasks with ChatGPT}


\author[1]{Andreas Scholl}{andreas.scholl@th-nuernberg.de}{0009-0009-1275-1374}
\author[2]{Daniel Schiffner}{d.schiffner@dipf.de}{0000-0002-0794-0359}
\author[3]{Natalie Kiesler}{natalie.kiesler@th-nuernberg.de}{0000-0002-6843-2729}

\affil[1]{Nuremberg Tech\\Computer Science\\Keßlerplatz 12, 90489 Nuremberg\\Germany}
\affil[2]{DIPF | Leibniz Institute for Research and Information in Education\\Information Centre Education\\Rostocker Straße 6, 60323 Frankfurt am Main\\Germany}
\affil[3]{Nuremberg Tech\\Computer Science\\Keßlerplatz 12, 90489 Nuremberg\\Germany}

\maketitle


\begin{abstract} 
Large Language Models (LLMs) have taken the world by storm, and students are assumed to use related tools at a great scale. In this research paper we aim to gain an understanding of how introductory programming students chat with LLMs and related tools, e.g., ChatGPT-3.5. To address this goal, computing students at a large German university were motivated to solve programming exercises with the assistance of ChatGPT as part of their weekly introductory course exercises. Then students (n=213) submitted their chat protocols (with 2335 prompts in sum) as data basis for this analysis. The data was analyzed w.r.t. the prompts, frequencies, the chats' progress, contents, and other use pattern, which revealed a great variety of interactions, both potentially supportive and concerning. Learning about students' interactions with ChatGPT will help inform and align teaching practices and instructions for future introductory programming courses in higher education.   
\end{abstract}


\section{Introduction} 

The increasing need for competent computing graduates proficient in programming, software development, and related technical competencies~\cite{cra_growth} is one of the factors exacerbating pressure on higher education institutions to offer high quality, competency-based education~\cite{raj2021fulldocument}. However, the latter requires extensive resources, mentoring, and, for example, formative feedback for learners, especially in introductory programming classes~\cite{jeuring2022towards,lohr2024feedback}. This is due to the fact that novices experience a number of challenges in the process, which have been subject to extensive research in the past decades~\cite{duboulay1986some,spohrer1986novice,Luxton-Reilly2018}. Among them are cognitively demanding competencies~\cite{kiesler2020towardsiticse,kiesler2024modeling}, such as problem understanding, designing and writing algorithms, debugging, and understanding error messages~\cite{kiesler2020towardsiticse,Luxton-Reilly2018,Ebert2016,spohrer1986novice,duboulay1986some}).
Educators' expectations towards novice learners and what they can achieve in their first semester(s) seem to be too high and unrealistic~\cite{luxton-reilly2016,Luxton-Reilly2018,whalley2007many}. 
Moreover, the student-educator ratio in introductory programming classes keeps increasing in German higher education institutions, thereby limiting resources to provide feedback and hints, and adequately address heterogeneous prior knowledge and diverse educational biographies~\cite{petersen2016revisiting,SB22}.

Fortunately, educational technologies based on Large Language Models (LLMs) have the potential to support novice learners of programming seeking help, if carefully used. LLMs and related tools have been shown to pass introductory programming tasks and courses~\cite{geng2023chatgpt,kiesler2023large,savelka2023large}, enhance programming error messages~\cite{Leinonen2023,Sarsa2022,macneil2022experiences,leinonen2023comparing}, and generate formative feedback~\cite{Bengtsson_Kaliff_2023,kiesler2023exploring,LMU-TEL/ADS2023,azaiz2024feedbackgeneration,roest2023nextstep} for learners. It is therefore not surprising that the emergence of LLMs triggered an extensive discussion within the computing education community regarding learning objectives, curricula, assessments and ethical questions~\cite{prather2023wgfullreport}. Due to the novelty of LLMs and their recent broad availability, most studies investigated LLMs from the educator's perspective, hypothesizing about (students') application scenarios (e.g.,~\cite{alhossami2024socratic,joshi2024chatgpt}). The student perspective (e.g., their trust or attitude towards LLMs~\cite{amoozadeh2024trust,rogers2024attitudes}), let alone their actual use of LLMs in a classroom setting, has only been addressed in a few very recent studies (e.g.,~\cite{liu2024teaching}). Understanding students' use of LLMs and respective interactions, however, is crucial to support them and adequately instruct them regarding the critical and reflective application of LLMs in their studies. 

Therefore, the present work investigates how students in an introductory programming class actually work with LLMs, such as ChatGPT, when doing their coursework. Specifically, it is guided by the research question (RQ): \textit{How do students chat with ChatGPT in the context of introductory programming course assignments?} 
To answer this question, an exercise sheet was developed as part of an introductory programming course at Goethe University in the winter term 2023/24. Students were instructed to solve programming tasks with the assistance of ChatGPT-3.5 (i.e., the freely available version at the time, December 2023), and to submit their chat protocols, which is the data basis of this work. The quantitative and qualitative analysis of the chat protocols provides the first insights of its kind into students' authentic interactions with the LLM in a curricular course setting. The contribution of this paper is thus a collection of students' interaction with an LLM, such as ChatGPT. These insights have implications for educators considering the use of LLMs in their introductory programming courses, as they can help inform the development of new forms of instructions with the goal of fostering a conscious and critical use.

\section{Related Work} 
\label{sec:relatedwork}

In 2019, a literature review of 146 articles on artificial intelligence applications in higher education~\cite{zawacki2019systematic} pointed out the rapidly increasing relevance of Artificial Intelligence (AI) in computing education and related fields. The review identified four main areas of interest: \enquote{adaptive systems and personalization, assessment and evaluation, profiling and prediction and intelligent tutoring systems} \cite[p.\,11]{zawacki2019systematic}. Notably, ~\citeauthor{zawacki2019systematic} recognized \enquote{the dramatic lack of critical reflection}~\cite[p.\,21]{zawacki2019systematic} of the challenges and risks of AI tools, including pedagogical and ethical considerations. 

Five years later, in 2024, Generative AI (GenAI) and related tools have already taken the world by storm~\cite{prather2023wgfullreport,prather2024wg}, and implications for higher education have become the focus of the discourse in many disciplines, including computing education research~\cite{macneil2024discussing}. This development is due to the broad availability of GenAI tools, which gained popularity with the launch of OpenAI's ChatGPT in late November 2022. Its impressive performance in, for example, introductory programming tasks and exams~\cite{geng2023chatgpt,kiesler2023large,savelka2023large} initiated discussions on LLMs' potential and how it would affect computing curricula, learning objectives, but also teaching, learning, and assessing in general~\cite{becker2023generative,prather2023wgfullreport,prather2023wg,kiesler2023beyond}. \citeauthor{prather2023wgfullreport} further emphasize the fast pace of GenAI tools and their developments, indicating the need for ongoing research to keep up with recent technological advances. This is especially true if research data (e.g., for benchmarking these tools) are not available~\cite{kiesler2023why,prather2023wgfullreport}.

General concerns and limitations of LLMs and GenAI tools comprise the accuracy and reliability in educational settings, as ChatGPT is susceptible to (re-)producing biased or inaccurate information~\cite{gill2024transformative}. The tool's knowledge base, which may lack recent information, can further lead to inaccuracies. Hallucinations add to this problem. Finally, it can bypass plagiarism detection tools. All of these aspects challenge the integrity of academic work and the reliability of AI as an educational tool, indicating the need for action by educational institutions~\cite{gill2024transformative,prather2023wgfullreport,zhai2022chatgpt}.

In the context of introductory programming education, studies revealed the potential of GenAI tools for several application scenarios. Among them are the effective generation of code explanations~\cite{macneil2022experiences,Sarsa2022}, enhanced programming error messages~\cite{Leinonen2023}, but also the analysis of student code with the goal of fixing students' errors~\cite{phung2023generating,zhang2022repairing}, or providing (elaborate) feedback~\cite{kiesler2023exploring,azaiz2024feedbackgeneration}. All of these studies focus on students' (potential) use of LLMs.  

For example, \citeauthor{macneil2022experiences} investigated students' perceptions towards automatically generated line-by-line code explanations by OpenAI's Codex and GPT-3 as part of an e-book. A majority of students evaluated the generated explanations as helpful~\cite{macneil2022experiences}. Similarly, \citeauthor{leinonen2023comparing} found that students rate code explanations generated by GPT-3 better on average than explanations from their peers. They also note that students showed no aversion towards LLM-generated feedback, and that student prefer line-by-line code explanations~\cite{leinonen2023comparing}. It is therefore not surprising that LLMs are being integrated into educational tools and environments to generate novice-friendly explanations tailored to each error~\cite{taylor2024dcchelperrorexplanations}, without revealing code solutions~\cite{kazemitabaar2024codeaid}, or to provide guardrails~\cite{liffiton2023codehelp}.

Related to that is the use of LLMs to analyze students' solutions and fix errors, or provide various types of elaborated feedback~\cite{narciss2006,shute} for novice programmers~\cite{keuning2018}. \citeauthor{phung2023generating} investigated the use of LLMs to fix syntax errors in Python programs and developed a technique to receive high precision feedback. Other studies qualitatively explored the feedback generated by ChatGPT~\cite{kiesler2023exploring,LMU-TEL/ADS2023,azaiz2024feedbackgeneration}. \citeauthor{kiesler2023exploring} characterized the feedback generated by ChatGPT-3.5 in response to authentic student solutions for introductory programming tasks. They found stylistic elements, textual explanations of the cause of errors and their fix, illustrating examples, meta-cognitive and motivational elements, but also misleading information, uncertainty in the model's response, and requests for more information by the LLM. \citeauthor{LMU-TEL/ADS2023}~\cite{LMU-TEL/ADS2023} noted difficulties of GPT-3.5 with the formatting of its output, recognizing correct solutions, and it hallucinated errors. \citeauthor{roest2023nextstep}~\cite{roest2023nextstep} conclude that the feedback generated by the GPT-3.5-turbo model seems to lack sufficient detail towards the end of an assignment. Nonetheless, a recent qualitative evaluation of GPT4 Turbo's feedback shows notable improvements, as the outputs are more structured, consistent, and always personalized~\cite{azaiz2024feedbackgeneration}.

While some of the aforementioned studies utilize authentic student data (e.g., students' solutions as an input to LLMs), others directly involve students to survey their perspective, or usability of LLMs. One example is the study by Prather et al.~\cite{prather2023s} who investigated students' use of GitHub Copilot in an introductory programming assignment. Their observations and interviews explore students' perceptions regarding the (usability) challenges and potential benefits of this technology, resulting in design implications. Similarly, \citeauthor{vaithilingam2022expectation} explored the usability of LLMs and their code generation abilities from the student perspective, concluding that Copilot's design should be improved~\cite{vaithilingam2022expectation}. 
\citeauthor{jayagopal2022exploring} explored the learnability of program synthesizers (including Copilot) by observing and interviewing students, which resulted in a set of lessons learned regarding system design. They also found the possibility to write a prompt to be perceived as more exciting and less confusing by students, compared to tools requiring format specifications~\cite{jayagopal2022exploring}. In another usability study, \citeauthor{barke2023grounded} presents a grounded theory analysis of programmers' interactions with Copilot, based on observations of 20 experienced programmers. 
They distinguish users in acceleration mode (i.e., Copilot is used to consciously achieve a goal faster) and exploration mode (i.e., Copilot is used to explore options). Moreover, they show that over-reliance on Copilot can reduce the task completion rate, and cognitive load can result from having to choose from multiple suggestions~\cite{barke2023grounded}.

In another study, \citeauthor{denny2023promptly} \cite{denny2023promptly} investigated first year students' interactions (n=54) with \textit{Promptly} in the context of a Python class. The tool utilizes the concept of \say{Prompt Problems}, aiming at teaching students how to prompt LLMs based on predefined problems. It automatically evaluates (i.e., executes and tests) code generated by an LLM (text-davinci-003 and gpt-3.5-turbo). In their analysis, however, \citeauthor{denny2023promptly} focus on the quantification of students' interactions with Promptly (e.g., number of words per prompt, prompts per problem) but not the follow-up interactions. Moreover, students' responses to reflective questions were used as a data basis. 

Even though many of these studies have investigated potential applications of LLMs by computing students, and specifically, novice programmers~\cite{leinonen2023comparing,macneil2022experiences,kiesler2023exploring,azaiz2024feedbackgeneration}, they are conducted from an educator's perspective, meaning all use cases had been predefined. So, they are not necessarily authentic. The presented usability studies~\cite{prather2023s,vaithilingam2022expectation,jayagopal2022exploring,barke2023grounded} were based on observations and interviews, and did not focus on novice programmers and their application of LLMs and related tools. The only study on students' interactions~\cite{denny2023promptly} emphasizes quantitative aspects, and does not analyze follow-up prompts and interaction patterns.

In addition, few very recent studies so far have been dedicated to computing students' trust in LLMs~\cite{amoozadeh2024trust}, and their attitude towards related tools~\cite{rogers2024attitudes}. Even fewer studies investigated students' use of LLMs based on actual interaction data. \citeauthor{liu2024teaching} investigated students' use of ChatGPT-3.5 as virtual teaching assistants in a classroom setting. However, they evaluated the tools' effectiveness (which was not based on students' interaction patterns)~\cite{liu2024teaching}. ~\citeauthor{grande2024studentperspective} evaluated the student perspectives on using ChatGPT for an assignment on professional ethics, but they used the students' submissions to the task as a data basis~\cite{grande2024studentperspective}. 
As a result, there is a lack of authentic interaction data of students using an LLM, i.e., their complete chat protocols, and their (qualitative) evaluation in the context of an introductory programming class.

\section{Methodology} 
\label{sec:meth}
To answer the RQ \textit{How do students chat with ChatGPT in the context of introductory programming course assignments?}, the study leverages empirical data collected from students (n=213) enrolled in an introductory programming course. Students were asked to complete a series of programming exercises with the assistance of ChatGPT-3.5, and to submit their chat protocols as part of the assignment. In this section, we introduce the course context, instructions, the selected tasks, and the methodology applied for data analysis. 

\subsection{Course Context Used for Data Gathering}
The context of this study was an introductory programming course for first-year computing students (n=790) at Goethe University Frankfurt (Germany) during the winter term 2023/24. The majority were computer science students, with some individuals from other majors who pursue computer science as a minor. The course is designed for novice learners or programmers, with no prerequisites. The class was accompanied by a Moodle course with learning materials. It comprised a 2-hour lecture per week for all 790 students, and a 2-hour tutorial session in groups of 20-30 students. A crucial components of the tutorial are the weekly or bi-weekly exercise sheets with programming tasks. Students are awarded points for (mostly individually) submitting their solutions, which contribute towards the exam held at the end of the semester. 

For the present study, a specific exercise sheet was developed for the tutorials in the week of December 6, 2023. Students were expected to submit their individual solutions two weeks later, group work was not permitted. Students were further instructed to \say{complete the tasks using ChatGPT via the free version} (3.5) on the web interface, and to submit \say{all prompts and responses} as paired entries in a spreadsheet via the Moodle course. 
However, no specific instructions were provided on how to interact with ChatGPT-3.5, except for a reference to OpenAI's guide on prompt engineering~\cite{openaiprompting}. This approach was deliberately chosen to avoid influencing students' interactions with ChatGPT-3.5.
The exercise sheet was available in both German and English to accommodate the diverse student body of the course. 
Using ChatGPT for this exercise sheet was voluntary, but two extra points were offered as an incentive. Moreover, students had been introduced to the present study's objective and the process during the lecture preceding the tutorial.

\subsection{Selected Tasks}

The exercise sheet addressed the concept of \textit{recursion}, \textit{functions}, \textit{lists}, \textit{conditionals}, \textit{string manipulation}, and \textit{documentation}. The exercise comprised two main tasks, while each one included sub-tasks. The first task, consisting of four sub-tasks, presented code snippets with recursive elements for several operations: (1a) summation of the digits of a number, (1b) reversing a list, (1c) performing multiplication, and (1d) computing the Ackermann function. Students were asked to read and interpret the given code snippets with the goal of determining the output of the code, the number of function calls, and identifying the implemented type of recursion.

The second task presented a more complex challenge, namely to implement a function that determines the number of \say{happy strings} within all sub-strings of a given string. A \say{happy string} was defined as a string that can either be rearranged into (or already is) a repetition of some string. To solve this task, students were required to (2a) test a given string for the \say{happy} property, and (2b) test all possible sub-strings of the given string. Students were encouraged to adopt a recursive approach by awarding additional points.

Prior to their integration into the exercise sheet, the authors evaluated ChatGPT-3.5's capability to solve the selected tasks. For task 1, ChatGPT demonstrated a high success rate in identifying the correct functions but frequently miscalculated the number of calls, often overlooking the base case. Task 2 proved to be more challenging for ChatGPT-3.5. Although the responses included the necessary steps for problem solving, it never included a correct solution (we tried 10 regenerations). This may have been due to a degree of ambiguity, and, for example, the uncommon term \say{happy string} in the task description.

\subsection{Data Analysis}

To evaluate students' interactions with ChatGPT-3.5, and how they seek assistance from the GenAI tool, the collected chat protocols were quantitatively and qualitatively analyzed. For the \textbf{quantification} of students' use patterns, the following aspects were investigated: 
\begin{itemize}
\itemsep2pt 
    \item \textit{Number of prompts per student}: The total count of prompts given to ChatGPT-3.5, offering insights into the extent of interaction from students with the GenAI tool for the given tasks. 
    \item \textit{Number of words per prompt}: Identifying the verbosity or conciseness of students' prompts, to help characterize students' descriptions of the problem or task (and its complexity).
    \item \textit{Follow-up interactions per student}: This characteristic aims to capture the iterative nature of problem-solving interactions with ChatGPT-3.5. It is supposed to help identify students' use patterns.
\end{itemize}

In addition, the chat protocols were \textbf{qualitatively} analyzed~\cite{mayring2000qualitative} by two of the authors to identify the type of follow-up interactions, the contents or issues addressed by the students' prompts, and the overall (interaction) strategy students seem to pursue or apply. 
\begin{itemize}
\itemsep2pt 
\item \textit{Solution requests (SR)}: Determines whether the student's prompt explicitly seeks a direct solution, shedding light on the intention behind the query and the expected outcome of the interaction (e.g., if only the task description is prompted). Frequencies emphasize the extend to which students apply this kind of request.
\item \textit{Type of follow-up interaction}: Categorizes whether a prompt is a standalone query (STA), a follow-up to a previous prompt (PRE), a response to a ChatGPT answer (RES), or a correction of a previous response (COR). We further add the frequencies of these types. Each prompt was categorized into only one of these four types.
\item \textit{Issue or problem solving step addressed by the prompt}: This classification identifies the issue addressed by the prompt. These categories include common problem solving steps and well-known issues for novice learners based on the literature~\cite{kiesler2020towardsiticse,Luxton-Reilly2018,Ebert2016,spohrer1986novice,duboulay1986some}: problem understanding (PU), conceptual understanding (CU), code generation (CG), debugging (DE), runtime analysis (RA), syntax/style (SY), documentation (DO), test cases (TC), and other categories (OT). It should be noted that a single prompt can reflect multiple issues within the problem solving process.
\item \textit{Categories describing students' interactions}: Describes how students navigate and modify their prompts based on the generated output. Categories reflecting these interaction pattern were inductively built based on the material.
\end{itemize}


\section{Results} 
A total of 360 students successfully completed the exercise sheet. Of these, 305 students engaged with ChatGPT-3.5 as part of the exercise. Due to the absence of the required template usage or submission in non-processable file formats, the data from 92 students could not be included in the final analysis. Therefore, the final dataset comprises 213 students, providing a collection of 2335 prompts for detailed examination. In total, we received 1668 German and 426 English prompts\footnote{\url{https://osf.io/WBKQV}}.



Students engaged in approximately 10.96 prompts on average. The median during all interactions is 7 prompts per student. Based on these numbers, we categorize students into three groups, as shown in \autoref{tab:quant_followup}: Group A, comprising individuals who use 0-5 prompts; Group B, including those who use 6-11 prompts; and Group C, consisting of students who use 12 or more prompts. The distinction yields almost equal sized groups.

\begin{table}
\centering
\footnotesize
\begin{tabular}{l|c|c|c|c||c|c|c|c}
 Range of prompts & No. of & No. of & Avg word & Solution & \multicolumn{4}{c}{Follow-Up Interactions} \\
per student & Students & Prompts & count & Requests (SR) & STA & PRE & RES & COR \\
 \hline
A: 0 to 5 & 66 & 262 & 73.74 & 213 & 172 & 57 & 25 & 8 \\
 \hline
B: 6 to 11 & 79 & 612 & 51.49 & 416 & 240 & 205 & 111 & 56 \\
 \hline
C: 12 and more & 68 & 1461 & 40.46 & 603 & 358 & 462 & 322 & 319 
\end{tabular}
\caption{Quantification of students' interactions with ChatGPT-3.5, including some of the qualitative aspects of their follow-up interactions.}
\label{tab:quant_followup}
\end{table}

Regarding the evaluation of word counts per prompt, it should be noted that the number of words for task 1 alone is 146, and 175 for task 2. This partially explains the decrease in the average word count from Group A to Group B by 30\%, followed by a further decline from Group B to Group C by 21\% to 40.46 words on average per prompt. Furthermore, the number of solution requests (SR) shows a declining trend across the groups. In Group A, approximately 81\% of prompts are SR, which decreases to 68\% in Group B and 41\% in Group C (see \autoref{tab:quant_followup}).

The nature of follow-up interactions (FIs) also varies among the groups (see \autoref{tab:quant_followup}). In Group A, the majority of interactions had no follow-up (STA, in 172 cases). In many cases, the prompts resembled the task descriptions, which had been used as input. At the same time, 90 instances of FIs were observed, comprising 57 follow-ups to a previous prompt (PRE), 25 responses to a generated answer by ChatGPT (RES), and 8 corrections of responses (COR), indicating a correction rate of approximately 3\%. 

Group B reveals an increase of FIs, with 372 instances of FIs, and somewhat fewer instances (240) without an FI (STA). 
Notably, there is a greater number of students referring back to their own prompts (PRE, 205 cases), while 56 corrections of a previous response (COR) were identified (9\%). 
In contrast, Group C demonstrates a significant increase of FIs, with the vast majority of follow-ups (1103 is sum). In Group 3, the correction rate (22\%) was particularly high, with 319 correction instances (COR).


The analysis of student prompts relating to programming problem-solving categories reveals several trends, summarized in \autoref{tab:problem_cat}.
For example, the proportion of initial prompts aimed at understanding the problem (PU), including solution requests, is smaller in longer conversations. In group A, these prompts constitute 76\% of the interactions, dropping to 50\% in group B, and further declining to 34\% in Group C.
Conversely, there is an increase of prompts requesting help in debugging (DE) in longer conversations. This category sees a rise from 6\% in group A to 17\% in group B, and a further increase to 24\% in group C.
The number of prompts referring to runtime analyses (RA) is higher for group A, accounting for 15\% of interactions. This may be due to fewer number of prompts in group A and runtime analysis being an explicit part of Task 1. 
In terms of concept understanding (CU), code generation (CG), and documentation (DO), there is a slight increase from group A to group B and C, while the percentage difference between group B and C is minimal.
Additionally, the analysis shows a relatively stable percentage of prompts related to syntax/style (SY) from group A to C, and a relatively consistent ratio for test case prompts (TC) to overall prompts observed across group A, B, and C (about 10\%).

\begin{table}[t!]
\centering
\begin{tabular}{l|c||c|c|c|c|c|c|c|c|c}
 Range of prompts & No. of & \multicolumn{9}{c}{Problem-solving category} \\
 \cline{3-11}
per student &  Prompts & PU & CU & CG & DE & RA & SY & DO & TC & OT  \\
 \hline
A: 0 to 5 & 262 & 198 & 25 & 34 & 16 & 40 & 3 & 24 & 26 & 3  \\
 \hline
B: 6 to 11 & 612 & 303 & 87 & 121 & 101 & 55 & 10 & 83 & 58 & 10  \\
 \hline
C: 12 and more & 1461 & 497 & 185 & 273 & 354 & 115 & 36 & 186 & 143 & 29 
\end{tabular}
\caption{Distribution of students' prompts across the various problem-solving categories (multiple categories per student and prompt).}
\label{tab:problem_cat}
\end{table}

After qualitatively analyzing the prompts, we were able to identify two common patterns of interactions with the GenAI tool: (1) the \textit{Task Description Prompts} pattern, and (2) the \textit{Prompts in Own Words} pattern. For (1), students seem to use the given task description at least as part of their initial prompt. In the other pattern (2), students rephrase the task description or immediately focus on specific parts of the problem statement. We provide a summary of both pattern and sub-pattern in \autoref{tab:ptd_examples} and \autoref{tab:pow_examples}, whereas the example ID refers to the student ID in the chat protocols. The order of the prompting patterns in the tables is supposed to reflect on the increasing specificity. 
\autoref{tab:ptd_examples} summarizes respective use pattern for \textit{Task Description Prompts}, showing different approaches to generate a solution using ChatGPT after having used the task description as an initial prompt.

\vspace{1em}
\begin{table}[h!]
\centering
\begin{tabular}{l|p{0.8\textwidth}}
\textbf{Example ID} & \textbf{Follow-ups after initial task description prompt:}\\
\hline
\hline
1012 & No or few additional instructions. \\
\hline
1044 & Giving direct orders / requests including style, documentation or corrections.\\
\hline
1199 & Short, but extensive prompting for explanations to gain understanding. \\
\hline
1135 & Specific prompts requesting explanations and corrections. \\
\hline
1070 & Testing of initially generated code by the LLM. Providing incorrect (console) output followed by \say{Correct the code} instructions. No additional input or adaptation.\\
\hline
1195 & Testing of initially generated code by the LLM. Providing incorrect (console) output followed by \say{Correct the code} instructions, resulting in disappointment.\\
\hline
1006 & Testing of initially generated code by the LLM. Providing incorrect (console) output followed by \say{Correct the code} instructions. Followed by instruction to restart, and adaptation of instruction.\\
\hline
1156 & Requesting (conceptual) explanations. Follow-up to create text and not bulletpoints.\\
\end{tabular}
\caption{Examples for the \textit{Task Description Prompts}-pattern encountered in the dataset.}
\label{tab:ptd_examples}
\end{table}

For the \textit{Prompt in Own Words} pattern (summarized in \autoref{tab:pow_examples}), students exhibit different approaches to achieve the correct solution or responses to their questions. These pattern seem more directed towards a specific aspect of the problem, or occurred only after the student had created a solution. 

\begin{table}[h!]
\centering
\begin{tabular}{l|p{0.8\textwidth}}
\textbf{Example ID} & \textbf{Prompts and follow-ups in students own words} \\
\hline
\hline
1014 & Demanding explanations for specific concepts, e.g. indices, dictionaries. \\
\hline
1058 & Adding own solution and asking specific questions.\\
\hline
1106 & Providing additional task constraints. Task description is provided as a follow up. Requests specific aspects, e.g., test cases, documentation, while providing examples. \\
\end{tabular}
\caption{Examples of \textit{Prompts in Own Words}-pattern found in the dataset.}
\label{tab:pow_examples}
\end{table}

\begin{figure}
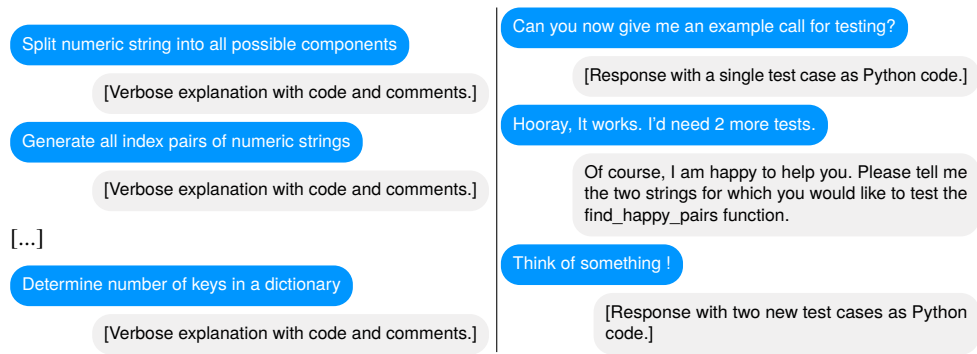

    \begin{subfigure}[b]{0.5\textwidth}
        \begin{prompt}
        \scriptsize
            Split numeric string into all possible components
        \end{prompt}
        \vspace{-2em}
        
        \begin{response}
        \scriptsize
            [Verbose explanation with code and comments.]
        \end{response}
        \vspace{-2em}
        
        \begin{prompt}
        \scriptsize
            Generate all index pairs of numeric strings
        \end{prompt}
        \vspace{-2em}
        
        \begin{response}
        \scriptsize
            [Verbose explanation with code and comments.]
        \end{response}
        \vspace{-0.5em}
        [...]
        \vspace{-0.5em}

        \begin{prompt}
        \scriptsize
            Determine number of keys in a dictionary
        \end{prompt}
        \vspace{-2em}
        
        \begin{response}
        \scriptsize
            [Verbose explanation with code and comments.]
        \end{response}

    \caption{Excerpt from chat protocol 1014 with brief and direct commands.}
    \label{fig:chat1037}
    \end{subfigure}
    \hfill
    \rulesep
    \begin{subfigure}[b]{0.5\textwidth}

        \begin{prompt}
        \scriptsize
            Can you now give me an example call for testing?
        \end{prompt}
        \vspace{-2em}
        
        \begin{response}
        \scriptsize
            [Response with a single test case as Python code.]
        \end{response}
        \vspace{-2em}
        
        \begin{prompt}
        \scriptsize
            Hooray, It works. I'd need 2 more tests.
        \end{prompt}
        \vspace{-2em}
        
        \begin{response}
        \scriptsize
            Of course, I am happy to help you. Please tell me the two strings for which you would like to test the find\_happy\_pairs function.
        \end{response}
        \vspace{-2em}

        \begin{prompt}
        \scriptsize
            Think of something !
        \end{prompt}
        \vspace{-2em}
        
        \begin{response}
        \scriptsize
            [Response with two new test cases as Python code.]
        \end{response}
    \caption{Excerpt from chat protocol 1037 resembling human-to-human interaction.}
    \label{fig:chat2}
    \end{subfigure}

\caption{Excerpts of two example chat protocols illustrating different interaction patterns (translated into English by the authors).}
\label{fig:chatprotocol1014}
\end{figure}

In addition to these two patterns, we observed some other interesting aspects of students' interactions. This includes switching the language from German to English as soon as the GenAI tool produced English outputs. Some prompts started in English, but students asked to create a documentation in German. Also, as part of the OT categorization, we found interjections, e.g., \say{Hooray}, \say{Thank You!}, which resemble human-to-human communication (ID 1037).
In \autoref{fig:chatprotocol1014}, we present two exemplary excerpts from students and their ChatGPT interactions. \autoref{fig:chat1037} illustrates that the student did not write complete sentences, but assigned explicit tasks to be executed. \autoref{fig:chat2} shows a  human-to-human interaction with the tool.

\vspace{1em}

\section{Discussion} 

The analysis of students' chat protocols indicates diverse interactions with GenAI tools, such as ChatGPT, when being applied as part of an introduction programming course in higher education. For example, students engaged differently with ChatGPT in terms of the number of prompts used, their length, and regarding (the content of) follow-up interactions.  

One group of students (n=66) only used 0-5 prompts while working on the selected tasks. This group may not have been thoroughly engaged in the interaction with ChatGPT. There is also a large number of solution requests (SR) present in this group. This may be an indicator for students looking for a quick solution to the given problems.

The second group of students (n=79) used ChatGPT differently. For example, we see an increasing number of prompts in this group. Still, many students requested the correct solution (SR). What is interesting though is that there are many follow-up interactions. This might indicate that either the initial response of ChatGPT was not considered good enough, or that the generated response actually fostered an interest in (an aspect of) the topic or problem. The increasing number of follow-ups on previous prompts (PRE) and responses to a ChatGPT answer (RES) disambiguate this group from students who only submitted 0-5 prompts.

The third group of students (n=68) used the GenAI tool most frequently, but for different purposes. For example, a smaller percentage of solution requests was identified. Especially due to the high number of corrections (COR), students seemed to be more reflective of the generated answers when trying to generate valid solutions. Especially interesting is the increase in responses to a ChatGPT answer (RES), which show that students actually used the tool as a \say{learning partner} instead of a simple query tool. This is also supported by the high numbers of prompts requesting help for understanding (PU) and debugging (DE).

The qualitative data analysis further reveals a difference in the categorization of problem-solving steps, the identified follow-up interactions, and prompting patters (\textit{Task Description Prompts} and \textit{Prompts in Own Words}). While the former are well known especially in the context of feedback generation~\cite{keuning2018,kiesler2023exploring}, the latter represent entirely new categories aiming to describe students' interactions and prompting pattern with ChatGPT. Educators can use these pattern as a basis for designing instructions on how to (not) use LLMs and related tools, and to reflect on their limitations and potential benefits. 

To answer our research question \textit{How do students chat with ChatGPT in the context of introductory programming course assignments}, we conclude that there is a certain variety and range of students' applications of the tool. There are at least three very different behaviors present. One seems to focus on problem solving by generating a single prompt to generate the best answer possible. Other students revealed a more \say{peer-like} interaction with the GenAI tool. They used more follow-up prompts, focused on problem solving step-by-step, tried to use the tool to identify and correct errors, and communicated respectively (we thus observed signs of anthropomorphism towards ChatGPT in this study). The third recognized use pattern is more straight forward and focuses on code generation, but utilizes the option of having an interactive tool. We assume that there are even more facets to these behaviors.


\section{Limitations}

A limitation of this study is the context of the data collection, which was a single class taking place at one university. Even though the high number of students participating in the study strengthens its representative character, the results may not be transferable to other educational contexts. Another limitation is that due to their knowledge of the study, students may have interacted differently with the tool than they would normally have in the context of an assignment, which is known as observer's paradox~\cite{roethlisberger1939management}.

\section{Conclusions and Future Work} 
This work presents the first-of-its kind exploratory study on students' use of ChatGPT as a tool to support problem-solving in the early stages of programming education at university level. To investigate students interactions, an exercise sheet had been designed and students were instructed to use ChatGPT, but not how to do so. Students' chat protocols were quantitatively and qualitatively analyzed to characterize students' use and interaction patterns. The high participation rate (n=213) allows us to draw several conclusions. For example, the results reveal three groups of students interacting differently with the tool. Among these groups are students seeking an immediate solution, but also students seeking help with problem understanding, conceptual knowledge, debugging, syntax error or style and documentation. Some students also used the chat more extensively than others, thereby resembling a \say{studdy buddy} or \say{virtual tutor}. The chat protocols further revealed new patterns of interactions, e.g., students starting with \textit{Task Description Prompts} and specifying their requests later, and those using \textit{Prompts in Own Words} with more specific requests. However, there is a huge diversity in students' use of ChatGPT. 

Implications for teaching and learning are manifold. For example, the diversity of use patterns reflects the heterogeneity of students' needs for support, and feedback. GenAI tools have the potential to offer that degree of personalization~\cite{azaiz2024feedbackgeneration}. Regardless, educators need to provide the necessary tools and advice on how to use them successfully, e.g., prompt patterns~\cite{white2023prompt}. Moreover, it is important to communicate the tools' limitations and potential pitfalls to students so they do not end up frustrated. The computing education community also needs to continue their discussion on the competencies required for successful graduates, and the role of GenAI tools in that process.

As our approach has been executed with a general LLM (ChatGPT-3.5), it would be interesting to see how a specialized AI tool for coding would increase its support for learners as part of future work. 
Moreover, a replication study at another institution or country would further strengthen the results. 
With the world facing new challenges and new tools every day, the potential in using AI for learning computer science is not yet fully understood. And to quote ChatGPT-3.5: \say{In this perpetual cycle of learning and teaching, computers enlighten students who, in turn, empower computers, perpetuating the beautiful recursion of knowledge.}

\printbibliography

@inproceedings{raj2021fulldocument,
author = {Raj, Rajendra and Sabin, Mihaela and Impagliazzo, John and Bowers, David and Daniels, Mats and Hermans, Felienne and Kiesler, Natalie and Kumar, Amruth N. and MacKellar, Bonnie and McCauley, Ren\'{e}e and Nabi, Syed Waqar and Oudshoorn, Michael},
title = {Professional Competencies in Computing Education: Pedagogies and Assessment},
year = {2021},
publisher = {ACM},
address = {New York},
doi = {10.1145/3502870.3506570},
booktitle = {Proceedings of the 2021 Working Group Reports on Innovation and Technology in Computer Science Education},
pages = {133–161},
numpages = {29},
}

@article{zawacki2019systematic,
  title={Systematic review of research on artificial intelligence applications in higher education--where are the educators?},
  author={Zawacki-Richter, Olaf and Marin, Victoria I and Bond, Melissa and Gouverneur, Franziska},
  journal={International Journal of Educational Technology in Higher Education},
  volume={16},
  number={1},
  pages={1--27},
  year={2019},
  publisher={Springer}
}

@misc{white2023prompt,
      title={A Prompt Pattern Catalog to Enhance Prompt Engineering with ChatGPT}, 
      author={Jules White and Quchen Fu and Sam Hays and Michael Sandborn and Carlos Olea and Henry Gilbert and Ashraf Elnashar and Jesse Spencer-Smith and Douglas C. Schmidt},
      year={2023},
      eprint={2302.11382},
      archivePrefix={arXiv},
      primaryClass={cs.SE}
}

@article{mayring2000qualitative,
	title={{Qualitative content analysis forum qualitative sozialforschung}},
	author={Mayring, Philipp},
	journal={{Forum: qualitative social research}},
	volume={1},
	number={2},
	year={2000}
}

@book{roethlisberger1939management,
	title={{Management and the Worker}},
	author={Roethlisberger, Fritz Jules and Dickson, William J.},
	year={1939},
	publisher={Harvard University Press},
	address={Cambridge}
}

@misc{denny2023promptly,
      title={Promptly: Using Prompt Problems to Teach Learners How to Effectively Utilize AI Code Generators}, 
      author={Paul Denny and Juho Leinonen and James Prather and Andrew Luxton-Reilly and Thezyrie Amarouche and Brett A. Becker and Brent N. Reeves},
      year={2023},
      eprint={2307.16364},
      archivePrefix={arXiv},
      primaryClass={cs.HC}
}

@article{becker2023generative,
  title={Generative AI in Introductory Programming},
  author={Becker, Brett A and Craig, Michelle and Denny, Paul and Keuning, Hieke and Kiesler, Natalie and Leinonen, Juho and Luxton-Reilly, Andrew and Prather, James and Quille, Keith},
  year={2023},
  url={https://csed.acm.org/wp-content/uploads/2023/12/Generative-AI-Nov-2023-Version.pdf},
publisher={ACM},

}

@article{zhai2022chatgpt,
	title={ChatGPT User Experience: Implications for Education},
	author={Xiaoming Zhai},
	doi={http://dx.doi.org/10.2139/ssrn.4312418},
	year={2022}
}

@inproceedings{taylor2024dcchelperrorexplanations,
author = {Taylor, Andrew and Vassar, Alexandra and Renzella, Jake and Pearce, Hammond},
title = {dcc --help: Transforming the Role of the Compiler by Generating Context-Aware Error Explanations with Large Language Models},
year = {2024},
publisher = {ACM},
address = {New York},
doi = {10.1145/3626252.3630822},
booktitle = {Proceedings of the 55th ACM Technical Symposium on Computer Science Education V. 1},
pages = {1314–1320},
numpages = {7},
}

@misc{phung2023generating,
      title={{Generating High-Precision Feedback for Programming Syntax Errors using Large Language Models}}, 
      author={Tung Phung and José Cambronero and Sumit Gulwani and Tobias Kohn and Rupak Majumdar and Adish Singla and Gustavo Soares},
      year={2023},
      eprint={2302.04662},
      archivePrefix={arXiv},
      primaryClass={cs.PL},
}

@inproceedings{kiesler2023why,
author = {Kiesler, Natalie and Schiffner, Daniel},
title = {{Why We Need Open Data in Computer Science Education Research}},
year = {2023},
doi = {10.1145/3587102.3588860},
publisher = {ACM},
address = {New York},
booktitle = {Proceedings of the 2023 Conference on Innovation and Technology in Computer Science Education V. 1},
pages = {348–353},
numpages = {6},
location = {Turku, Finland},
series = {ITiCSE 2023}
}

@misc{kiesler2023beyond,
  title={Beyond the Textbook: Rethinking Students' Competencies in the LLM Era},
  author={Natalie Kiesler},
  howpublished={Generative AI: Implications for Teaching and Learning},
  address={Uppsala, Sweden},
  doi={10.13140/RG.2.2.28355.37922/1},
  year={2023}
}

@inproceedings{grande2024studentperspective,
author = {Virginia Grande and Natalie Kiesler and Maria Andreina Francisco Rodriguez },
title = {{Student Perspectives on Using a Large Language Model (LLM) for
an Assignment on Professional Ethics}},
year = {2024},
doi = {10.1145/3649217.3653624},
publisher = {ACM},
address = {New York},
booktitle = {Proceedings of the 2024 Conference on Innovation and Technology in Computer Science Education V. 2},
location = {Milan, Italy},
series = {ITiCSE 2024}
}

@inproceedings{lohr2024feedback,
author = {Lohr, Dominic and Kiesler, Natalie and Keuning, Hieke and Jeuring, Johan},
title = {{\say{Let Them Try to Figure It Out First} – Reasons Why Experts (Do Not) Provide Feedback to Novice Programmers}},
year = {2024},
doi = {10.1145/3649217.3653530},
publisher = {ACM},
address = {New York},
booktitle = {Proceedings of the 2024 Conference on Innovation and Technology in Computer Science Education V. 1},
location = {Milan, Italy},
series = {ITiCSE 2024}
}

@article{zhang2022repairing,
  title={Repairing Bugs in Python Assignments Using Large Language Models},
  author={Zhang, Jialu and Cambronero, Jos{\'e} and Gulwani, Sumit and Le, Vu and Piskac, Ruzica and Soares, Gustavo and Verbruggen, Gust},
  journal={arXiv preprint arXiv:2209.14876},
  year={2022}
}

@article{gill2024transformative,
    title = {Transformative effects of ChatGPT on modern education: Emerging Era of AI Chatbots},
    journal = {Internet of Things and Cyber-Physical Systems},
    volume = {4},
    pages = {19-23},
    year = {2024},
    doi = {10.1016/j.iotcps.2023.06.002},
    author = {Sukhpal Singh Gill and Minxian Xu and Panos Patros and Huaming Wu and Rupinder Kaur and Kamalpreet Kaur and Stephanie Fuller and Manmeet Singh and Priyansh Arora and Ajith Kumar Parlikad and Vlado Stankovski and Ajith Abraham and Soumya K. Ghosh and Hanan Lutfiyya and Salil S. Kanhere and Rami Bahsoon and Omer Rana and Schahram Dustdar and Rizos Sakellariou and Steve Uhlig and Rajkumar Buyya}
}

@inproceedings{liffiton2023codehelp,
author = {Liffiton, Mark and Sheese, Brad E and Savelka, Jaromir and Denny, Paul},
title = {CodeHelp: Using Large Language Models with Guardrails for Scalable Support in Programming Classes},
year = {2024},
publisher = {ACM},
address = {New York},
doi = {10.1145/3631802.3631830},
booktitle = {Proceedings of the 23rd Koli Calling International Conference on Computing Education Research},
articleno = {8},
numpages = {11},
location = {Koli, Finland},
series = {Koli Calling '23}
}

@article{duboulay1986some,
	title={{Some difficulties of learning to program}},
	author={Du Boulay, Benedict},
	journal={{Journal of Educational Computing Research}},
	volume={2},
	number={1},
	pages={57--73},
	year={1986},
    doi = {10.2190/3LFX-9RRF-67T8-UVK9},
}

@article{whalley2007many,
	title={{The many ways of the Bracelet project}},
	author={Whalley, Jacqueline and Clear, Tony and Lister, Raymond},
	journal={{BACIT}},
	year={2007},
	publisher={National Advisory Committee on Computing Qualifications},
}

@inproceedings{Luxton-Reilly2018,
	author = {Luxton-Reilly, Andrew and Simon and Albluwi, Ibrahim and Becker, Brett A. and Giannakos, Michail and Kumar, Amruth N. and Ott, Linda and Paterson, James and Scott, Michael James and Sheard, Judy and Szabo, Claudia},
	title = {{Introductory Programming: A Systematic Literature Review}},
	year = {2018},
	publisher = {ACM},
	address = {New York},
	booktitle = {{Proc. ITiCSE}},
	pages = {55--106},
	numpages = {52},
    doi = {10.1145/3293881.3295779}
}

@inproceedings{petersen2016revisiting,
  title={Revisiting why students drop CS1},
  author={Petersen, Andrew and Craig, Michelle and Campbell, Jennifer and Tafliovich, Anya},
  booktitle={Proc. Koli Calling},
  pages={71--80},
  year={2016},
  doi={10.1145/2999541.2999552}
}

@inproceedings{kiesler2020towardsiticse,
author = {Natalie Kiesler},
title = {{Towards a Competence Model for the Novice Programmer Using Bloom's Revised Taxonomy -- An Empirical Approach}},
year = {2020},
publisher = {ACM},
address = {New York},
doi = {10.1145/3341525.3387419},
booktitle = {{Proceedings of the 2020 ACM Conference on Innovation and Technology in Computer Science Education}},
pages = {459--465},
numpages = {7},
location = {Trondheim, Norway},
series = {ITiCSE '20}
}

@inproceedings{luxton-reilly2016,
	author = {Luxton-Reilly, Andrew},
	title = {{Learning to Program is Easy}},
	year = {2016},
	doi = {10.1145/2899415.2899432},
	booktitle = {{Proc. ITiCSE}},
	pages = {284--289},
	numpages = {6},
}

@article{spohrer1986novice,
	title={{Novice mistakes: Are the folk wisdoms correct?}},
	author={Spohrer, James C. and Soloway, Elliot},
	journal={{Communications of the ACM}},
	volume={29},
	number={7},
	pages={624--632},
	year={1986},
	publisher={ACM New York},
    doi = {10.1145/6138.6145}
}

@inproceedings{Ebert2016,
  title={A presentation framework for programming in programing lectures},
  author={Ebert, Michael and Ring, Markus},
  booktitle={Proc. EDUCON},
  pages={369--374},
  year={2016},
  organization={IEEE}
}

@misc{geng2023chatgpt,
      title={{Can ChatGPT Pass An Introductory Level Functional Language Programming Course?}}, 
      author={Chuqin Geng and Yihan Zhang and Brigitte Pientka and Xujie Si},
      year={2023},
      eprint={2305.02230},
      archivePrefix={arXiv},
      primaryClass={cs.CY},
}

@Article{LMU-TEL/ADS2023,
  author  = {Azaiz, Imen and Deckarm, Oliver and Strickroth, Sven},
  journal = {International Journal of Engineering Pedagogy (iJEP)},
  title   = {AI-enhanced Auto-Correction of Programming Exercises: How Effective is GPT-3.5?},
  year    = {2023},
  month   = dec,
  number  = {8},
  pages   = {67--83},
  volume  = {13},
  doi     = {10.3991/ijep.v13i8.45621},
}

@misc{roest2023nextstep,
      title={Next-Step Hint Generation for Introductory Programming Using Large Language Models}, 
      author={Lianne Roest and Hieke Keuning and Johan Jeuring},
      year={2023},
      eprint={2312.10055},
      archivePrefix={arXiv},
      primaryClass={cs.CY}
}

@misc{savelka2023large,
      title={Large Language Models (GPT) Struggle to Answer Multiple-Choice Questions about Code}, 
      author={Jaromir Savelka and Arav Agarwal and Christopher Bogart and Majd Sakr},
      year={2023},
      eprint={2303.08033},
      archivePrefix={arXiv},
      primaryClass={cs.CL}
}

@misc{Bengtsson_Kaliff_2023, 
series={TRITA-EECS-EX}, title={Assessment Accuracy of a Large Language Model on Programming Assignments}, 
url={https://urn.kb.se/resolve?urn=urn:nbn:se:kth:diva-331000}, abstractNote={The education sector is changing rapidly and adopting new practices of managing student assignments. Manually assessing student work can be costly, as well as sometimes erroneous, implying it can be beneficial to automate the grading process. With new large language models rises questions on how well these can be used to grade assignments, as well as how their accuracy can be improved. This study has explored how task context affects the accuracy of a large language model when grading programming assignments. The large language model used in this study was OpenAI’s recently released GPT-4 model, and the student assignments were collected from an introductory programming course (DD1338) at KTH Royal Institute of Technology. In order to evaluate the grading accuracy, this study first had to inject errors into correct student assignments, which was also done with GPT-4. Four different logical error categories were used: looping, if-else, recursion, and off-by-one errors. When the large language model was provided with the instruction context, the results were sometimes inconsistent but indicated that task context negatively affects the assessment accuracy. In addition, the feedback provided in the assessment seem to hold a high accuracy level. Even though the feedback may become more accurate when the model is provided with the instruction context, this usually comes with fewer identified errors overall and therefore a smaller assessment accuracy. Several recommendations for future research are recommended, including investigating how other types of context impact the accuracy, or how well the large language model identifies other types of errors. Future studies might also investigate how grading the same file several times might affect the accuracy, due to the non-deterministic nature of a large language model such as GPT-4.}, 
author={Bengtsson, Douglas and Kaliff, Axel}, 
year={2023}, 
collection={TRITA-EECS-EX} 
}

@inproceedings{macneil2022experiences,
author = {MacNeil, Stephen and Tran, Andrew and Hellas, Arto and Kim, Joanne and Sarsa, Sami and Denny, Paul and Bernstein, Seth and Leinonen, Juho},
title = {{Experiences from Using Code Explanations Generated by Large Language Models in a Web Software Development E-Book}},
year = {2023},
doi = {10.1145/3545945.3569785},
booktitle = {Proc. SIGCSE TS},
pages = {931–937},
numpages = {7},
}

@inproceedings{leinonen2023comparing,
author = {Leinonen, Juho and Denny, Paul and MacNeil, Stephen and Sarsa, Sami and Bernstein, Seth and Kim, Joanne and Tran, Andrew and Hellas, Arto},
title = {{Comparing Code Explanations Created by Students and Large Language Models}},
year = {2023},
doi = {10.1145/3587102.3588785},
booktitle = {Proc. ITiCSE},
pages = {124--130},
}

@inproceedings{alhossami2024socratic,
author = {Al-Hossami, Erfan and Bunescu, Razvan and Smith, Justin and Teehan, Ryan},
title = {Can Language Models Employ the Socratic Method? Experiments with Code Debugging},
year = {2024},
publisher = {ACM},
address = {New York},
doi = {10.1145/3626252.3630799},
booktitle = {Proceedings of the 55th ACM Technical Symposium on Computer Science Education V. 1},
pages = {53–59},
numpages = {7},
location = {Portland, OR, USA, },
series = {SIGCSE 2024}
}

@inproceedings{joshi2024chatgpt,
author = {Joshi, Ishika and Budhiraja, Ritvik and Dev, Harshal and Kadia, Jahnvi and Ataullah, Mohammad Osama and Mitra, Sayan and Akolekar, Harshal D. and Kumar, Dhruv},
title = {ChatGPT in the Classroom: An Analysis of Its Strengths and Weaknesses for Solving Undergraduate Computer Science Questions},
year = {2024},
publisher = {ACM},
address = {New York},
doi = {10.1145/3626252.3630803},
booktitle = {Proc. of the 55th ACM Technical Symposium on Computer Science Education V. 1},
pages = {625–631},
numpages = {7},
}

@inproceedings{liu2024teaching,
author = {Liu, Rongxin and Zenke, Carter and Liu, Charlie and Holmes, Andrew and Thornton, Patrick and Malan, David J.},
title = {Teaching CS50 with AI: Leveraging Generative Artificial Intelligence in Computer Science Education},
year = {2024},
publisher = {ACM},
address = {New York},
doi = {10.1145/3626252.3630938},
booktitle = {Proceedings of the 55th ACM Technical Symposium on Computer Science Education V. 1},
pages = {750–756},
series = {SIGCSE 2024}
}

@inproceedings{macneil2024discussing,
author = {MacNeil, Stephen and Leinonen, Juho and Denny, Paul and Kiesler, Natalie and Hellas, Arto and Prather, James and Becker, Brett A. and Wermelinger, Michel and Reid, Karen},
title = {Discussing the Changing Landscape of Generative AI in Computing Education},
year = {2024},
publisher = {ACM},
address = {New York},
doi = {10.1145/3626253.3635369},
booktitle = {Proceedings of the 55th ACM Technical Symposium on Computer Science Education V. 2},
pages = {1916},
}

@inproceedings{rogers2024attitudes,
author = {Rogers, Michael P. and Hillberg, Hannah Miller and Groves, Christopher L.},
title = {Attitudes Towards the Use (and Misuse) of ChatGPT: A Preliminary Study},
year = {2024},
publisher = {ACM},
address = {New York},
doi = {10.1145/3626252.3630784},
booktitle = {Proceedings of the 55th ACM Technical Symposium on Computer Science Education V. 1},
pages = {1147–1153},
numpages = {7},
location = {Portland, OR, USA, },
series = {SIGCSE 2024}
}

@misc{kiesler2023large,
      title={Large Language Models in Introductory Programming Education: ChatGPT's Performance and Implications for Assessments}, 
      author={Natalie Kiesler and Daniel Schiffner},
      year={2023},
      eprint={2308.08572},
      doi={10.48550/arXiv.2308.08572},
      archivePrefix={arXiv},
      primaryClass={cs.SE}
}

@inproceedings{amoozadeh2024trust,
author = {Amoozadeh, Matin and Daniels, David and Nam, Daye and Kumar, Aayush and Chen, Stella and Hilton, Michael and Srinivasa Ragavan, Sruti and Alipour, Mohammad Amin},
title = {Trust in Generative AI among Students: An exploratory study},
year = {2024},
publisher = {ACM},
address = {New York},
doi = {10.1145/3626252.3630842},
booktitle = {Proceedings of the 55th ACM Technical Symposium on Computer Science Education V. 1},
pages = {67–73},
numpages = {7},
}

@inproceedings{prather2023wg,
author = {Prather, James and Denny, Paul and Leinonen, Juho and Becker, Brett A. and Albluwi, Ibrahim and Caspersen, Michael E. and Craig, Michelle and Keuning, Hieke and Kiesler, Natalie and Kohn, Tobias and Luxton-Reilly, Andrew and MacNeil, Stephen and Petersen, Andrew and Pettit, Raymond and Reeves, Brent N. and Savelka, Jaromir},
title = {Transformed by Transformers: Navigating the AI Coding Revolution for Computing Education: An ITiCSE Working Group Conducted by Humans},
year = {2023},
publisher = {ACM},
address = {New York},
doi = {10.1145/3587103.3594206},
booktitle = {Proceedings of the 2023 Conference on Innovation and Technology in Computer Science Education V. 2},
pages = {561–562},
numpages = {2},
}

@inproceedings{prather2024wg,
author = {Prather, James and Leinonen, Juho and Kiesler, Natalie and Gorson Benario, Jamie and Lau, Sam and MacNeil, Stephen and Norouzi, Narges and Opel, Simone and Pettit, Virginia and Porter, Leo and Reeves, Brent and Savelka, Jaromir and Smith IV, David H. and Strickroth, Sven and Zingaro, Daniel},
title = {How Instructors Incorporate Generative AI into Teaching Computing},
year = {2024},
publisher = {ACM},
address = {New York},
doi = {10.1145/3649405.3659534},
booktitle = {Proceedings of the 2024 Conference on Innovation and Technology in Computer Science Education V. 2},
location = {Milan, Italy},
series = {ITiCSE 2024}
}

@inproceedings{prather2023wgfullreport,
author = {Prather, James and Denny, Paul and Leinonen, Juho and Becker, Brett A. and Albluwi, Ibrahim and Craig, Michelle and Keuning, Hieke and Kiesler, Natalie and Kohn, Tobias and Luxton-Reilly, Andrew and MacNeil, Stephen and Petersen, Andrew and Pettit, Raymond and Reeves, Brent N. and Savelka, Jaromir},
title = {The Robots Are Here: Navigating the Generative AI Revolution in Computing Education},
year = {2023},
publisher = {ACM},
address = {New York},
doi = {10.1145/3623762.3633499},
booktitle = {Proceedings of the 2023 Working Group Reports on Innovation and Technology in Computer Science Education},
pages = {108–159},
}

@article{barke2023grounded,
author = {Barke, Shraddha and James, Michael B. and Polikarpova, Nadia},
title = {Grounded Copilot: How Programmers Interact with Code-Generating Models},
year = {2023},
publisher = {ACM},
address = {New York},
volume = {7},
number = {OOPSLA1},
doi = {10.1145/3586030},
journal = {Proc. ACM Program. Lang.},
month = {4},
articleno = {78},
numpages = {27}
}

@inproceedings{vaithilingam2022expectation,
	title = {Expectation vs. Experience: Evaluating the Usability of Code Generation Tools Powered by Large Language Models},
	shorttitle = {Expectation vs. Experience},
	pages = {1--7},
booktitle = {},	
 bbooktitle = {{CHI} Conference on Human Factors in Computing Systems Extended Abstracts},
	author = {Vaithilingam, Priyan and Zhang, Tianyi and Glassman, Elena L.},
	year = {2022},
	publisher = {ACM},
address = {New York},
}

@inproceedings{jayagopal2022exploring,
  title={Exploring the learnability of program synthesizers by novice programmers},
  author={Jayagopal, Dhanya and Lubin, Justin and Chasins, Sarah E},
  booktitle={Proceedings of the 35th Annual ACM Symposium on User Interface Software and Technology},
  pages={1--15},
  year={2022}
}

@article{kazemitabaar2024codeaid,
  title={CodeAid: Evaluating a Classroom Deployment of an LLM-based Programming Assistant that Balances Student and Educator Needs},
  author={Kazemitabaar, Majeed and Ye, Runlong and Wang, Xiaoning and Henley, Austin Z and Denny, Paul and Craig, Michelle and Grossman, Tovi},
  journal={arXiv preprint arXiv:2401.11314},
  year={2024}
}

@article{prather2023s,
author = {Prather, James and Reeves, Brent N. and Denny, Paul and Becker, Brett A. and Leinonen, Juho and Luxton-Reilly, Andrew and Powell, Garrett and Finnie-Ansley, James and Santos, Eddie Antonio},
title = {“It’s Weird That It Knows What I Want”: Usability and Interactions with Copilot for Novice Programmers},
year = {2023},
doi = {10.1145/3617367},
journal = {ACM Trans. Comput.-Hum. Interact.},
}

@article{keuning2018,
author = {Keuning, Hieke and Jeuring, Johan and Heeren, Bastiaan},
title = {A Systematic Literature Review of Automated Feedback Generation for Programming Exercises},
year = {2018},
issue_date = {March 2019},
publisher = {ACM},
address = {New York},
volume = {19},
number = {1},
doi = {10.1145/3231711},
journal = {ACM Trans. Comput. Educ.},
month = {9},
articleno = {3},
numpages = {43}
}

@book{narciss2006,
	title = {Informatives Tutorielles Feedback: Entwicklungs- und Evaluationsprinzipien auf der Basis instruktionspsychologischer Erkenntnisse},
	author = {Narciss, Susanne},
	year = {2006},
	publisher = {Waxmann Verlag},
	address={M\"{u}nster}
}

@Article{shute,
  author  = {Shute, Valerie J.},
  title   = {{Focus on formative feedback}},
  journal = {Review of Educational Research},
  year    = {2008},
  volume  = {78},
  number  = {1},
}

@InProceedings{Sarsa2022,
  author     = {Sarsa, Sami and Denny, Paul and Hellas, Arto and Leinonen, Juho},
  booktitle  = {Proc. ICER},
  title      = {Automatic Generation of Programming Exercises and Code Explanations Using Large Language Models},
  year       = {2022},
  month      = aug,
  publisher  = {ACM},
  doi        = {10.1145/3501385.3543957},
}

@InProceedings{Leinonen2023,
  author     = {Leinonen, Juho and Hellas, Arto and Sarsa, Sami and Reeves, Brent and Denny, Paul and Prather, James and Becker, Brett A.},
  booktitle  = {Proc. SIGCSE},
  title      = {Using Large Language Models to Enhance Programming Error Messages},
  year       = {2023},
  month      = mar,
  publisher  = {ACM},
  doi        = {10.1145/3545945.3569770},
}

@InProceedings{SB22,
  author       = {Strickroth, Sven and Bry, François},
  booktitle    = {Proc. CSEDU},
  title        = {{The Future of Higher Education is Social and Personalized! Experience Report and Perspectives}},
  year         = {2022},
  pages        = {389--396},
  volume       = {1},
  doi          = {10.5220/0011087700003182},
  project      = {gate},
}

@book{kiesler2024modeling,
author="Kiesler, Natalie",
title="Modeling Programming Competency: A Qualitative Analysis",
year="2024",
publisher="Springer International Publishing",
address="Cham",
pages="165",
isbn="978-3-031-47148-3",
doi="10.1007/978-3-031-47148-3",
url="https://doi.org/10.1007/978-3-031-47148-3"
}

@inproceedings{kiesler2023exploring,
  author={Kiesler, Natalie and Lohr, Dominic and Keuning, Hieke},
  booktitle={2023 IEEE Frontiers in Education Conference (FIE)}, 
  title={Exploring the Potential of Large Language Models to Generate Formative Programming Feedback}, 
  year={2024},
  volume={},
  number={},
  pages={1-5},
  doi={10.1109/FIE58773.2023.10343457}
}

@misc{azaiz2024feedbackgeneration,
      title={Feedback-Generation for Programming Exercises With GPT-4}, 
      author={Imen Azaiz and Natalie Kiesler and Sven Strickroth},
      year={2024},
      eprint={2403.04449},
      archivePrefix={arXiv},
      primaryClass={cs.AI},
      doi={10.48550/arXiv.2403.04449}
}

@misc{openaiprompting,
  author = {OpenAI},
  year = {2023},
  title = {Prompt Engineering},
  howpublished = {\url{https://platform.openai.com/docs/guides/prompt-engineering}},
  note = {Accessed: 2024-04-08}
}

@misc{cra_growth,
    author = {Camp, Tracy and Adrion, W. Richards and Bizot, Betsy and Davidson, Susan and Hall, Mary and Hambrusch, Susanne and Walker, Ellen and Zweben, Stuart},
    title = {Generation CS: the growth of computer science},
    year = {2017},
    issue_date = {June 2017},
    publisher = {ACM},
    address = {New York},
    volume = {8},
    number = {2},
    issn = {2153-2184},
    doi = {10.1145/3084362},
    journal = {ACM Inroads},
    month = {5},
    pages = {44–50},
    numpages = {7}
}

@inproceedings{jeuring2022towards,
author = {Jeuring, Johan and Keuning, Hieke and Marwan, Samiha and Bouvier, Dennis and Izu, Cruz and Kiesler, Natalie and Lehtinen, Teemu and Lohr, Dominic and Peterson, Andrew and Sarsa, Sami},
title = {Towards Giving Timely Formative Feedback and Hints to Novice Programmers},
year = {2022},
publisher = {ACM},
address = {New York},
doi = {10.1145/3571785.3574124},
booktitle = {Proceedings of the 2022 Working Group Reports on Innovation and Technology in Computer Science Education},
pages = {95–115},
numpages = {21},
location = {Dublin, Ireland},
series = {ITiCSE-WGR '22}
}

\end{document}